# Generational Differences in Automobility: Comparing America's Millennials and Gen Xers Using Gradient Boosting Decision Trees

Kailai Wang[a], Xize Wang[b, *]

**Abstract**

Whether the Millennials are less auto-centric than the previous generations has been widely discussed in the literature. Most existing studies use regression models and assume that all factors are linear-additive in contributing to the young adults' driving behaviors. This study relaxes this assumption by applying a non-parametric statistical learning method, namely the gradient boosting decision trees (GBDT). Using U.S. nationwide travel surveys for 2001 and 2017, this study examines the non-linear dose-response effects of lifecycle, socio-demographic and residential factors on daily driving distances of Millennial and Gen-X young adults. Holding all other factors constant, Millennial young adults had shorter predicted daily driving distances than their Gen-X counterparts. Besides, residential and economic factors explain around 50% of young adults' daily driving distances, while the collective contributions for life course events and demographics are about 33%. This study also identifies the density ranges for formulating effective land use policies aiming at reducing automobile travel demand.

**Keywords:** Millennials; Life course events; Gradient Boosting Decision Trees; Driving distance; VMT; Machine Learning.

a. Department of Construction Management, University of Houston, Houston, TX, USA. Email: kwang43@cougarnet.uh.edu. OCRID: 0000-0002-5597-6823.
b. Department of Real Estate, National University of Singapore, Singapore. Email: wangxize316@gmail.com. OCRID: 0000-0002-4861-6002
* Corresponding author.

**Introduction**

Millennials' travel behaviors have drawn widespread attention over the last decade (e,g, McDonald, 2015; Blumenberg et al., 2016; Garikapati et al., 2016; Delbosc et al., 2019). Broadly speaking, Millennials refer to those born within the final two decades of the 20th century. Millennials are the first generation to have grown up surrounded by digital devices and Internet connectivity. They have also experienced the 2007-2009 Great Recession when entering adulthood (Klein and Smart, 2017; Manville et al., 2017). These social and economic trends may have shaped them to hold different lifestyles and travel choices than earlier generations at the same age level (Lee and Circella, 2019). As driving has been the dominant travel mode in the U.S. for decades, an improved understanding of Millennials' automobility, such as how and why Millennials' automobility differs from prior generations, can offer insights for policy makers to plan for future-ready cities.

Empirical evidence shows that the Millennials drive less frequently and in shorter distances than earlier generations; however, whether such differences will persist as Millennials age remains an ongoing debate (Blumenberg et al., 2016; McDonald, 2015). On one hand, the "demographic perspective" sees the generational differences in attitudes towards automobility as a by-product of the experiences in early life stages and may be constant over time. For instance, Smart and Klein (2018) find that if a person is exposed to high-quality public transit services during young adulthood, this experience affects his or her travel behavior in later life stages. On the other hand, the "life course perspective" argues that the generational differences in automobility may be due to different "schedules" in reaching life milestones (e.g., being employed, getting married, etc.) for people born in different time periods (Delbosc and Nakanishi, 2017; Scheiner and Holz-Rau, 2013; Susilo et al., 2019). Thereby, the observed difference may disappear when the younger generation "catch up" with the delayed life course schedules. For instance, the Millennial generation delay in hitting the milestones of adulthood compared with previous generations – partially due to the lingering effects of the global economic recession in 2008 (Blumenberg et al., 2016).

Some recent studies have fully explored the young adults sample of the 2017 U.S. National Household Travel Survey (NHTS) and compared their driving patterns with those at the same age in previous nationwide travel surveys (e.g., da Silva et al., 2019; Knittel and Murphy, 2019; Wang, 2019; Wang and Akar, 2020). These studies control for various confounders, including socio-demographic characteristics, life course events, adoption of technology advances, and residential location choices. In addition to cross-generational comparisons controlling for these "confounding" factors, an equally important question is the relative contributions of these "contributing" factors to young adults' automobility, which has not been extensively discussed yet. A closer look at each factor's contribution to automobility can provide useful information to policy makers and professionals for updating policy measures. From a methodological perspective, machine learning models can complement the regression-based studies by estimating each factor's relative contribution without assuming that these factors contribute to automobility in a linear-additive manner. Applying machine learning methods can also help to examine the non-linear dose-response relationship between each factor and automobility, which could help planners to develop more nuanced land use and transport plans.

By adopting a recently-emerging machine learning approach, gradient boosting decision trees (GBDT), this study analyzes America's Millennials and Gen Xers' driving patterns at 21-36 years old. This study's dataset comes from the 2001 and 2017 U.S. National Household Travel Surveys (NHTSs) (Federal Highway Administration, 2004, 2018a). Specifically, our GBDT model estimates the relative contributions of residential location choice, economic conditions, life course events, information and communication technology (ICT) usages, and demographics to Millennials and Gen Xers' daily driving distances. Furthermore, this study estimates the non-linear dose-response effects of each factor for Millennial and Gen-X young adults. We compared Millennials and Gen Xers' expected daily driving distances under various hypothetical scenarios. These scenarios assume different combinations of economic conditions, lifecycle stages, and residential location choices; the between-generational differences are thus assumed to be explained by generational differences in other factors, such as attitudes and preferences.

**Literature Review**

Recent empirical evidence shows that a higher proportion of today's young adults in the U.S. live in compact, transit-oriented communities when entering adulthood than previous generations (Myers, 2016; Talen, 2017). They have pro-environmental attitudes, sustainable travel behaviors (i.e., active travel, taking mass transit and using shared-mobility services), and rely on the use of ICT for daily activities (e.g., e-commerce or social media) (Circella et al., 2016; Hopkins, 2016; van Wee, 2015; Wang et al., 2018). This phenomenon has also been emerging beyond the U.S. In many developed countries, this generation's young adults are observed to take longer time to acquire a driving license, own fewer vehicles, and drive less (Kuhnimhof et al., 2012; Delbosc and Currie, 2013; McDonald, 2015; Hjorthol, 2016; Klein and Smart, 2017; Thigpen and Handy, 2018; Wang et al., 2018; Delbosc et al., 2019). These generational differences may reflect the shifts in beliefs and values, moving away from the "car culture" to "multimodality" (e.g., Circella et al., 2016). If such a trend persists and will not lead to lower accessibility to opportunities, it will help planners and policy makers reshape the transportation systems to be more sustainable (Ralph, 2017).

Young adults' changing mobility patterns are a composite outcome of various explanatory factors, including macroeconomic conditions, socio-demographic characteristics, life course events, residential location choices, and adoption of technology advances. The economic recession during 2008-2009 had much stronger impacts on Millennials than other generations. When entering early adulthood, Millennials had to experience harsh labor and housing markets (Furman, 2014; Mawhorter, 2018); they had lower employment rates, reduced incomes, and increased gas prices (Blumenberg et al., 2016; Klein and Smart, 2017; Manville et al., 2017). According to media reports, there are concerns that Millennials' economic adversary might be carried over to later life stages (e.g., Luhby, 2018). The financial hardships also made them delay marriage and childbearing (Garikapati et al., 2016).

To quantify each factor's contribution to young adults' automobility, most studies utilize regression models including ordinary least squares, binary models, and count models (Blumenberg et al., 2016; Choi et al., 2017; da Silva et al., 2019; Knittel and Murphy, 2019; McDonald, 2015; Wang, 2019; Wang and Akar, 2020; Zhong and Lee, 2017). For example, using data from the 1989-2014 Puget Sound household travel surveys, Zhong and Lee (2017) investigated the factors that influence the decline of vehicle ownership among young adults since the mid-2000s in the Puget Sound region, Washington State. They estimated a Poisson model to bridge the links among demographic factors, urban forms, and the number of vehicles per household. Zhong and Lee (2017) tested the shifts in attitudes among young adults by adding interaction terms created between the main variables of interest and survey years. A projection was then conducted based on the regression estimates. The results reveal that the effect of Millennial-specific factors (i.e., attitudinal shifts in location and mobility preferences) is smaller than that of demographic characteristics on the reduction in vehicle ownership.

Besides regression models, data-driven, non-linear models have been emerging in the transportation planning literature. Recently, there has been rich literature on applying machine learning approaches to predict travel demand and provide utility interpretations. These approaches include naïve Bayes (Ma et al., 2013), tree-based ensemble methods (Zhang and Haghani, 2015; Ding et al., 2018a), neural networks (Xie et al., 2003), and support vector machines (Pirra and Diana, 2019). As one of the tree-based ensemble methods, the GBDT algorithm was argued to provide better predictions than regression-based approaches (Zhang and Haghani, 2015; Ma et al., 2017). Recent studies also show great interest in exploring the utility interpretation of using GBDT (e.g., Ding et al., 2018a; Ding et al., 2018b; Dong et al., 2019; Tao et al., 2020). These works show that GBDT is powerful in capturing non-linear effects and relative contributions of explanatory factors, particularly when high-level interaction terms exist. As one of the first studies utilizing this approach, Ding et al. (2018a) used data from the Oslo metropolitan area, Norway to examine the association between the built environment and driving distance. They find that distance from one's residence to the city center has the largest effect among all built environment variables tested. They also find that most built environment variables tend to show non-linear relationships with driving distance, indicating that the built environment – travel associations are stronger within specific value ranges.

Tree-based machine learning models could contribute to the literature on generational differences in mobility in two ways. First, by relaxing the linear-additive assumptions, tree-based machine learning models assume that different factors can have more complex interactions than linear-additive relationships. In regression models, context-specific effects such as gender-specific effects of childbearing on automobility are usually measured as interaction terms between gender and having children. However, three-way or higher-order interaction terms (e.g., gender-specific impacts of childbearing on automobility for workers and non-workers in transit-oriented vs. sprawled neighborhoods) are unusual for regression models due to the difficulty in estimation and interpretation. Such high-order interactions may or may not exist; nevertheless, machine learning methods do not directly rule out these complex interaction effects, and hence would help estimate the relative contributions of various factors on young adults' mobility patterns. Second, machine-learning models can examine the non-linear dose-response relationships between contributing factors (e.g., residential density) and automobility. Such a non-linear relationship can complement the regression-based studies that commonly report the average effects. For instance, the non-linear dose-response relationships can help recognize if the marginal effects are larger in specific value ranges of the explanatory variables. Such methods can also help in examining whether dose-response relationships differ across generations. However, we need to caution that tree-based machine learning models cannot make statistical inferences, such as levels of significance and confidence intervals which are the strengths of regression-based models.

Specifically, this study applies the gradient boosting decision tree (GDBT) approach and aims to make the two contributions mentioned above. This study estimates the relative contributions of lifestyle, socio-demographic, and residential factors to Millennial and Gen X young adults' daily driving distances without the linear-additive assumptions. This study also examines the non-linear dose-response effects of various explanatory factors for Millennials and Gen Xers.

**Data and Methods**

*Data source and study sample*

The individual- and household-level data for this study come from the 2001 and 2017 National Household Travel Surveys (NHTSs), conducted by the U.S. Federal Highway Administration (2004, 2018a). Each survey includes information on individuals, households, and vehicles, as well as detailed travel diaries on a survey day. Note that the 2001 and 2017 NHTSs have different sampling schemes: the 2001 survey is based on the Random Digit Dialing (RDD) frame and only includes households with landline phones; the 2017 survey is based on address-based sampling and includes households with and without landline phones (e.g., cellphone-only households) (Federal Highway Administration, 2004, 2018a). The 2017 survey has a larger sample size than the 2001 survey: the number of households surveyed for the 2017 NHTS is 129,696, while the number for the 2001 NHTS is 66,038. The sample size difference between the two surveys is due to the fact that a higher number of state and local governments participated the Add-on Program of the 2017 NHTS[1]. As the NHTS public data does not have identifiers on which observations belong of the "original" national sample and which observations belong to the add-ons, we use all observations in our analysis. For descriptive statistics, tabulations and t-tests, we adjust for personal weights to ensure that the 2001 and 2017 samples are both nationally-representative; for the GBDT model, we followed earlier studies in the literature (e.g. McDonald, 2015) and did not adjust for personal weights, as unweighted data is preferable if factors used in the development of weights are directly included in the models (Winship and Radbill, 1994).

---

[1] The NHTS Add-on Program allows states and metropolitan planning organizations to purchase additional samples in NHTS to support state- or metropolitan-level planning and forecasting. For a detailed introduction, see: https://nhts.ornl.gov/addOn.shtml

Following the Pew Research Center (2018), we define Millennials as those born between 1981 and 1996; and the preceding generation, Gen Xers, as those born between 1965 and 1980. Millennials were 21-36 years old in 2017; and Gen Xers were at the same age in 2001. Hence, the dataset of this study includes young adults between the ages of 21 and 36 from the 2001 NTHS (Gen Xers) and the 2017 NHTS (Millennials). To focus on the drivers' behaviors, the study sample excludes young adults without driver's licenses. According to the weighted NHTS samples, 88.8% of the Gen Xers and 91.8% of the Millennials hold drivers' licenses. As per existing studies, the 3% generational difference should be explained largely by the observed generational differences in their attitudes towards driving and their levels of "traveling impendence" given by parents (McDonald et al., 2016; Thigpen and Handy, 2018). The study sample also excludes young adults who drove more than 300 miles on the survey day (the 99.5th percentile of the daily driving distance). The final study sample includes 53,912 young adults from 21 to 36 years old: 20,678 of them is from the 2001 survey and belong to Gen X, and 33,234 of them is from the 2017 survey and are Millennials. The higher share of the 2017 young adults in the study sample reflects the larger overall sample size of the 2017 NHTS.

The outcome variable of this study is driving distance on the survey day (in miles); for mixed-mode trips such as "Park and Ride", the driving segment will be included in the daily driving mileages. Distance of driving, or vehicle mileage, is regarded as a close proxy for vehicle greenhouse gas emissions by the literature (e.g. Boarnet et al., 2017). The explanatory variables include residential location, economic conditions, life course events, ICT usages, demographics and time-specific indicators. Following earlier studies analyzing NHTS data (Blumenberg et al., 2016; McDonald, 2015), we transformed the categorical variables on household income and residential density as continuous variables (see Table 1 for details). We then adjusted for inflation factors to convert the income data in 2001 into 2017 US Dollars[2].

Note that as the technology advances, the 2017 NHTS calculated the trip distances using an online mapping interface, while the 2001 NHTS uses self-reported trip distances. To make the two surveys comparable, we have inflated the 2017 trip distances by 10%, as suggested by Federal Highway Administration (2018b). We admit that the 2001 and 2017 driving distance variables may still be incomparable even after the adjustments. As a robustness check, we re-ran our model using another driving behavior variable, daily auto trip frequency, as the outcome variable. We used automobile trip frequency since it is not impacted by the different distance-calculation methods between the two surveys. The outputs of this auto trip frequency model still support our main conclusions on the generational differences in automobility, and are available upon request.

Table 1 shows the descriptive statistics for this study sample (adjusted for survey weights), as well as the results of two-sample t-tests between the 2001 and 2017 sub-sample. As Table 1 shows, the 2001 and 2017 young adults significantly differ in all variables except for gender shares and economic conditions. Among the driver's license holders, young adults drove for fewer miles in 2017 as compared to their counterparts in 2001 (32.0 miles vs. 35.6 miles). A higher proportion of Millennials have bachelor's or higher degrees than Gen Xers, which corresponds with other national-level surveys such as American Community Survey and Current Population Survey (U.S. Census Bureau, 2019). The Millennials are more internet-savvy than their Gen-X counterparts during young adulthood: higher shares of the 2017 young adults telecommute and use internet on a daily basis. Millennials also have a relatively delayed life cycle: compared with their Gen-X counterparts, they are less likely to have children and more likely to live with parents or other older relatives. A higher percentage of young adults in 2017 choose to live in relatively high-density areas than their counterparts in 2001, implying that Millennial young adults prefer denser, urban neighborhoods compared their Gen-X counterparts at the age of 21-36 (Lee, 2018). Also, the young adults in 2017 were relatively less likely to be homeowners, which is in line with earlier studies using NHTS. We do not observe significant differences between two survey years on two factors representing economic conditions: number of household vehicles per person and household incomes, implying that the U.S. economy has gradually recovered from the Great Recession (DePillis, 2017).

---

[2] Source of inflation adjustment factors: https://www.usinflationcalculator.com/

## Table 1. Data Summary for the Young Adults (21-36 years old) Sample [a]

| Survey year | 2001 | | 2017 | | p-value for t-test or chi-square tests |
|---|---|---|---|---|---|
| | Mean/Percent | S.D. | Mean/Percent | S.D. | |
| ***Outcome Variable*** | | | | | |
| Driving Distance on the survey day | 35.55 | 40.06 | 31.97 | 40.40 | <0.001 |
| ***Residential Location*** | | | | | |
| Population density (1,000 persons per square mile) [b] | 5.18 | 6.47 | 6.18 | 7.49 | <0.001 |
| Size of metropolitan areas | | | | | <0.001 |
|    Not residing in MSAs | 16.2% | | 11.5% | | |
|    In an MSA of less than 1,000,000 | 23.7% | | 31.0% | | |
|    In an MSA or CMSA of 1,000,000 - 2,999,999 | 22.8% | | 21.7% | | |
|    In an MSA or CMSA of 3 million or more | 37.3% | | 35.8% | | |
| Neighborhood type [c] | | | | | <0.001 |
|    Urban | 15.4% | | 22.6% | | |
|    Others (suburban, second city, small town, rural) | 84.6% | | 77.4% | | |
| ***Economic Conditions*** | | | | | |
| Number of vehicles per person in household | 0.77 | 0.45 | 0.78 | 0.43 | 0.233 |
| Household incomes (in 1,000 2017 US$) [d] | 79.35 | 54.77 | 79.63 | 59.67 | 0.780 |
| ***Life Course Events*** | | | | | |
| Household lifecycle | | | | | <0.001 |
|    Has young children (15 or younger) | 56.1% | | 43.5% | | |
|    Does not have young children | 43.9% | | 56.5% | | |
| Formed household | | | | | <0.001 |
|    Lives independently | 72.6% | | 64.9% | | |
|    Lives with parents or other older relatives | 27.4% | | 35.1% | | |
| Homeownership | | | | | <0.001 |
|    Owner | 60.7% | | 51.4% | | |
|    Non-owner | 39.3% | | 48.6% | | |
| Household size | 3.39 | 1.48 | 3.18 | 1.42 | <0.001 |
| ***ICT Effects*** | | | | | |
| Frequency of telecommute | | | | | <0.001 |
|    Telecommute weekly | 2.8% | | 5.2% | | |
|    Telecommute monthly | 2.0% | | 9.8% | | |
|    Telecommute is available option | 1.3% | | 2.7% | | |
|    Cannot telecommute | 93.9% | | 82.3% | | |
| Access to internet | | | | | <0.001 |
|    Daily web user | 43.8% | | 97.6% | | |
|    Weekly web user | 17.8% | | 1.3% | | |
|    Monthly web user | 10.3% | | 0.3% | | |
|    Use web less than monthly | 28.2% | | 0.8% | | |
| ***Socio-Demographics*** | | | | | |
| Age | 29.16 | 4.48 | 28.90 | 4.64 | <0.001 |
| Gender | | | | | 0.948 |
|    Male | 50.2% | | 50.2% | | |
|    Female | 49.8% | | 49.8% | | |

| | | | |
|---|---|---|---|
| Education | | | <0.001 |
|     Less than high school | 6.7% | 2.5% | |
|     High school graduate | 27.0% | 14.7% | |
|     Some college or associate degree | 32.2% | 31.4% | |
|     Bachelor's or higher degree | 34.1% | 51.4% | |
| Race of household head | | | 0.002 |
|     Non-Hispanic White | 68.2% | 62.5% | |
|     Non-Hispanic Black | 10.2% | 9.1% | |
|     Non-Hispanic Asian/Pacific Islander | 3.4% | 6.1% | |
|     Hispanic | 15.7% | 17.9% | |
|     Other races | 2.6% | 4.4% | |
| Employment Status | | | <0.001 |
|     Worker | 86.7% | 83.7% | |
|     Non-worker | 13.3% | 16.3% | |
| ***Time-Specific Indicators*** | | | |
| Survey day | | | <0.001 |
|     Weekday | 69.2% | 73.7% | |
|     Weekend | 30.8% | 26.4% | |
| Number of observations | 20678 | 33234 | |

Notes:
  a. The proportions are reported for categorical variables; the mean values with standard deviations are reported for continuous variables. The summary statistics are adjusted by survey weights.
  b. NHTS reports population density using a list of discrete intervals. The author constructs a continuous indictor for this factor based on the midpoint of each category. For example, if the population density of a respondent is reported as "2,000-3,999 persons per square mile", this study takes the value as "3000 persons per square mile". For the density category (≥25,000 persons per square mile), this study uses 30,000 persons per square mile.
  c. This variable is based on the NHTS Urban/Rural indicator. The indicator categorizes census tracts into five community types: urban, suburban, second city, small town and rural. The categorization is based on density and locations. For details, see Claritas (2018).
  d. Similar to the population density, the author takes the values at the midpoints of the reported categories. If any individual reports the annual household incomes as "$5,000 – $9,999", we take the value as "$7,500". For respondents from the 2001 and 2017 NHTSs, we measure the highest interval "≥ $100,000" as "150,000"; for respondents from the 2017 NHTS, we measure the highest interval "≥$200,000" as "250,000". As the buying power of the dollar changes over time, the incomes of 2001 are transformed into 2017 US dollars (website: https://www.usinflationcalculator.com/).

*The GDBT model*

This study applies the gradient boosting decision tree (GBDT) model to investigate the relative contributions of explanatory variables to young adults' daily driving distances, as well as to examine the different non-linear dose-response effects for the explanatory variables for Millennials (young adults in 2017) and Gen Xers (young adults in 2001). The GBDT algorithm begins with classifying the whole sample into sub-groups, and the predicted values for each sub-group's observations are estimated by multiple linear regressions. Then the generated errors will be used to adjust each independent variable's weight in the next round of prediction. During this iteration, a set of decision trees is built. The GBDT algorithm calculates the prediction error at each stage of the iteration. The prediction errors determine the optimal number of trees. Researchers usually use the model with a minimized prediction error for further interpretations. Compared with the linear regression models, the GBDT method can generate more accurate predictions and handle variables with missing values. More importantly, in our case, utilizing the GBDT approach can deal with the uncertain interactions among predictors. For instance, the effects of childbearing on daily driving distance may differ by gender, employment status, and neighborhood type at the same time. Regression models can only accurately capture such effects by assuming four-way interaction terms among (a) having children, (b) gender, (c) worker status, and (d) urban neighborhood indicator, while GBDT models do not need such strong assumptions when making estimations. Future details of the comparative advantages of GBDT models over linear models are documented in Chung (2013), Ding et al. (2018a), Ding et al. (2018b), Dong et al. (2019), and Zhang and Haghani (2015).

This study adopts a five-fold cross-validation procedure to develop the GBDT model. In other words, the algorithm randomly shuffled the dataset into five distinct subsets and then picked four subsets (80%) to train the model and used the remaining subset (20%) to validate; the algorithm repeated this process five times whereas each subset served as validation subset for once, and the average was taken as the final results. There are three key parameters to consider when estimating the GBDT model: number of trees, learning rate, and tree complexity (Friedman, 2001). Following the previous travel behavior studies using the GBDT model (e.g., Chung, 2013; Ding et al., 2018a; Ding et al., 2018b), this study sets a maximum of 15,000 trees and chooses the learning rate at 0.001. As to tree complexity, this parameter reflects the independent variables' actual interactions (Friedman, 2001). This study estimates a series of models by increasing the interaction depth level of trees from one to twenty. A complexity of 15 is finally selected since it reaches a balance between the goodness of fit and the potential overfitting problem. We fitted the GBDT models using the *"gbm"* R package (Greenwell et al., 2019) and estimated the marginal effects (i.e., partial dependence) at different points using the *"pdp"* R package (Greenwell, 2018).

**Results**

After 10,569 boosting iterations, we have identified the final GDBT model based on the value of squared error loss. This section shows the non-linear dose-response effects for explanatory variables, and their relative contributions to young adults' daily driving distances.

*Non-linear effects of key explanatory variables*

To investigate the non-linear dose-response effect of the explanatory variables on daily driving distances for Gen Xers and Millennials, this section presents the partial dependence plots of the GDBT model. The partial dependence plots estimate the dependent variables' values for the explanatory variable at different intervals while holding other variables at their mean values. In each plot, young adults in 2001 (Gen Xers) and 2017 (Millennials) are illustrated separately.

The three partial dependent plots in Figure 2(a-c) illustrate the associations between key explanatory variables and daily driving distance. As indicated in Figure 2(a), residential density is negatively associated with daily driving distances for both Millennials and Gen Xers. For both generations, the associations between residential density and daily driving distances are almost linear at the range of 0–6,000 persons per square mile; beyond this range, the associations tend to be much flatter. Figure 2(a) also indicated that holding all other factors constant, Millennials have higher expected daily driving distances than their Gen-Xer counterparts across different levels of population density. Admittedly, Millennials' relatively lower automobility levels can be attributed to their somewhat higher concentration in dense urban neighborhoods (see Table 1); nevertheless, this finding shows that Millennial young adults drive less than their Gen X counterparts, even after holding neighborhood density constant. The results are consistent with earlier studies based on linear models (e.g., Wang, 2019).

Figure 2(b) shows that, for both Millennials and Gen Xers, annual household income has an overall positive association with daily driving distance. Their driving distances increase substantially with annual household incomes, when annual household incomes is lower than approximately 100,000 USD (in 2017 US Dollars). After that, the income-automobility association becomes relatively flat. Furthermore, the gaps in predicted driving distances between Gen Xers and Millennials are almost constant across the entire range. In addition, holding other factors constant, the predicted daily driving distances for Millennial young adults are consistently lower than their Gen-X counterparts across all income levels. Such findings may imply a shift in car culture for today's higher-income young adults, as suggested by an earlier study that adjusts annual household incomes by household size (i.e., Wang and Akar, 2020).

As shown in Figure 2(c), when the number of household vehicles per person is lower than two, both Gen Xers' and Millennials' daily driving distance increases substantially with average vehicle numbers. In contrast, the total daily driving distance does not vary much when per capita household vehicle ownership is more than two. The overall results suggest that Millennials' daily driving distances are consistently lower than that for their Gen X counterparts, while the two generations' dose-response plots are in similar shapes.

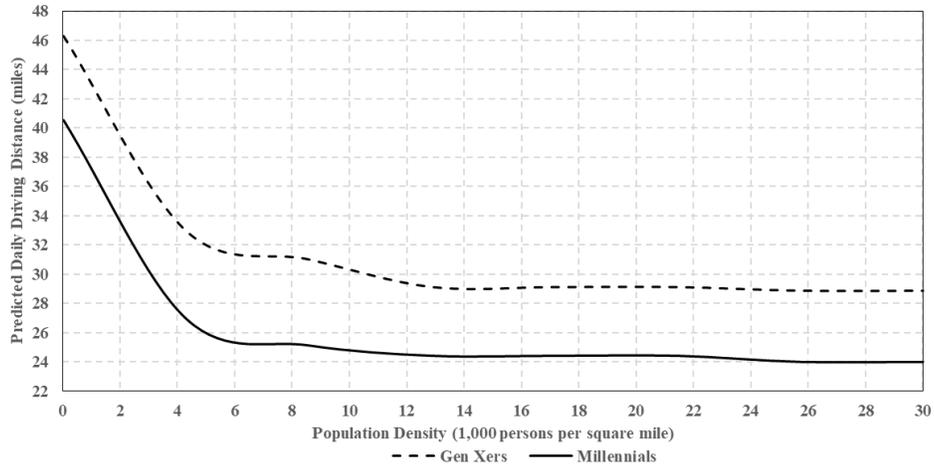
(a) population density

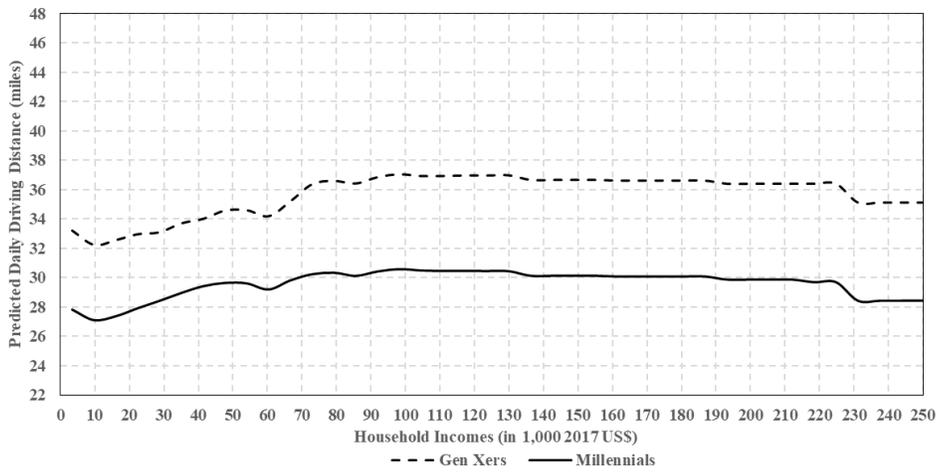
(b) household incomes

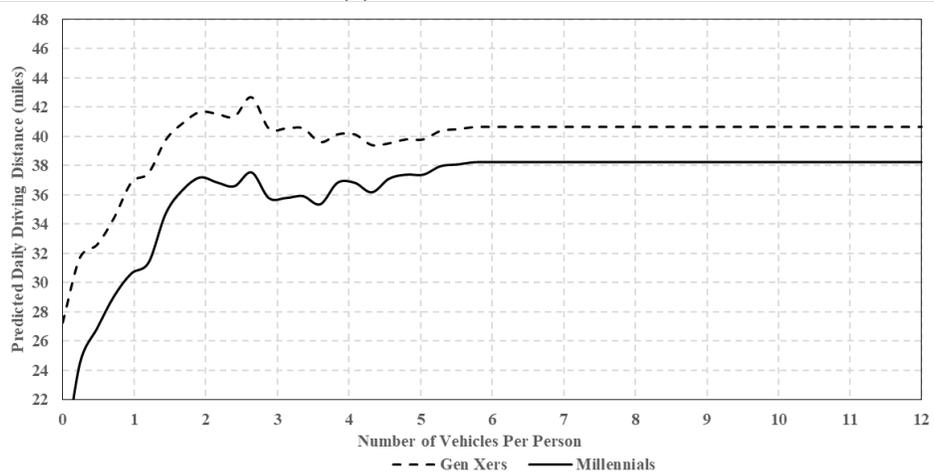
(c) number of vehicles per person in household

**Figure 2. Partial dependent plots of residential and economic factors on daily driving distances**

As a comparison, Table 2 tabulates the average daily driving distances for the 2001 and 2017 young adults sample by the three factors shown in Figure 2. Comparing Table 2 with Figure 2, we can see that Millennial young adults drive less than their Gen-X counterparts in different levels of population density, household income and number of vehicles per person in household, both with and without adjustments for life course events, ICT effects and other socio-demographic factors. Such findings show factors such as delayed life courses among the Millennials can only partially explains their relative lower driving distances than their Gen-X counterparts. In other words, even controlling for delayed life-courses, Millennials still drive less than Gen Xers.

**Table 2. Average daily driving distances for 2001 and 2017 young adults, by residential and economic factors**

|  | 2001 | | 2017 | | Difference between 2001 and 2017 |
|---|---|---|---|---|---|
|  | Obs. | Driving Distance | Obs. | Driving Distance |  |
| **Population density (1000 persons per square mile)** | | | | | |
| 0-100 | 2809 | 46.95 | 2964 | 39.44 | -7.51 |
| 100-500 | 3193 | 40.69 | 4222 | 33.67 | -7.03 |
| 500-1,000 | 1835 | 37.25 | 2984 | 32.87 | -4.38 |
| 1,000-2,000 | 2612 | 36.48 | 4281 | 28.68 | -7.80 |
| 2,000-4,000 | 3559 | 37.59 | 6411 | 26.53 | -11.06 |
| 4,000-10,000 | 4767 | 33.68 | 8735 | 24.06 | -9.62 |
| 10,000-25,000 | 1395 | 26.29 | 2783 | 21.40 | -4.89 |
| 25,000- | 508 | 26.30 | 854 | 13.95 | -12.34 |
| **Household incomes (in 1,000 2017 US$)** | | | | | |
| Less than $10,000 | 222 | 23.81 | 1182 | 18.38 | -5.43 |
| $10,000 to $14,999 | 495 | 29.87 | 924 | 19.18 | -10.69 |
| $15,000 to $24,999 | 1381 | 31.83 | 1970 | 23.59 | -8.24 |
| $25,000 to $34,999 | 871 | 34.27 | 2657 | 24.83 | -9.43 |
| $35,000 to $49,999 | 2625 | 32.65 | 4044 | 29.30 | -3.35 |
| $50,000 to $74,999 | 5681 | 36.83 | 6388 | 29.78 | -7.05 |
| $75,000 to $99,999 | 3728 | 39.65 | 5526 | 28.47 | -11.17 |
| $100,000 to $149,999 | 3548 | 40.02 | 6479 | 29.33 | -10.69 |
| $150,000 or more | 2127 | 38.63 | 4064 | 25.95 | -12.69 |
| **Number of vehicles per person in household** | | | | | |
| Less than 0.5 | 3318 | 32.85 | 4020 | 18.54 | -14.31 |
| Between 0.5 to 1 | 8683 | 35.09 | 11344 | 27.29 | -7.80 |
| between 1 to 2 | 8019 | 39.36 | 16651 | 29.91 | -9.45 |
| between 2 to 3 | 561 | 42.51 | 977 | 40.90 | -1.61 |
| More than 3 | 97 | 29.41 | 242 | 29.47 | 0.06 |

Note: the average daily driving distances are adjusted by the NHTS personal weights.

Table 3 demonstrates the associations between life course events and daily driving distances. Due to the complexity of the multiple interactions, Table 3 reports the predicted daily driving distances with different value combinations. Holding all other variables constant, the 2017 young adults had lower predicted daily driving distances than their 2001 counterparts across all the eight possible combinations. For instance, holding all other factors constant, the predicted daily driving distance for Millennial young adults who have children, live independently, and own the houses is 31.3 miles, while their Gen-X counterparts is expected to drive 5.4 miles longer. Depending on specific combinations, the difference ranges from 5.2 to 6.1 miles, indicating that Millennial young adults drive less than their Gen X counterparts when in the same lifecycle conditions.

**Table 3. The Marginal Effects of Life Course Events and Survey Years**

| Have Children[a] | Live Independently[a] | Own home[a] | Predicted Daily Driving Distance (miles) | | |
|---|---|---|---|---|---|
| | | | 2001 | 2017 | Difference (2017-2001) |
| 0 | 0 | 0 | 32.94 | 27.09 | -5.86 |
| 0 | 0 | 1 | 33.05 | 27.17 | -5.88 |
| 0 | 1 | 0 | 34.08 | 28.00 | -6.07 |
| 0 | 1 | 1 | 34.20 | 28.10 | -6.10 |
| 1 | 0 | 0 | 35.46 | 30.22 | -5.24 |
| 1 | 0 | 1 | 35.58 | 30.32 | -5.26 |
| 1 | 1 | 0 | 36.59 | 31.16 | -5.43 |
| 1 | 1 | 1 | 36.71 | 31.27 | -5.44 |

Note: a: "1" = "yes"; "0" = "no".

Figure 3 illustrates the marginal effects of ICT usage on young adults' driving distances. The frequency of telecommuting is positively associated with daily driving distance for both 2001 and 2017 young adults. The result implies that policy provisions aimed at promoting telecommuting may increase car dependency among young people. This study also finds that the actual contribution of internet usage on driving distance is small, as estimated marginal effects are almost constant and do not vary much across different levels of internet usage.

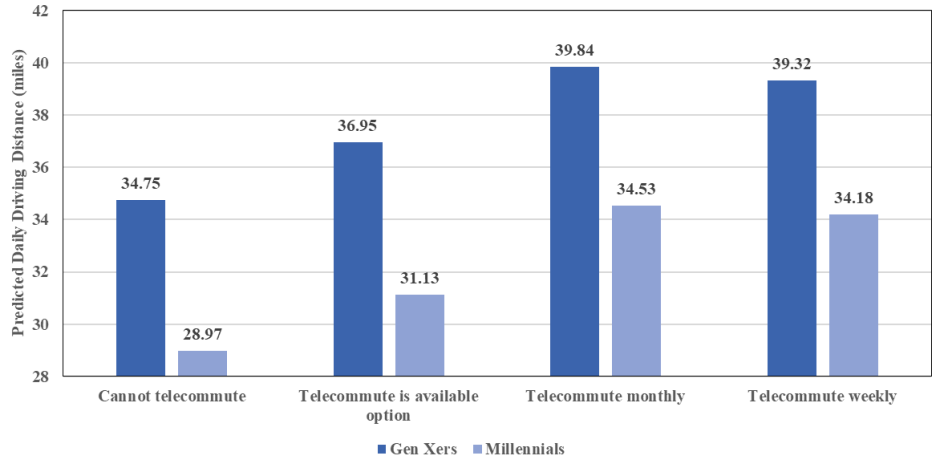

(a) telecommuting

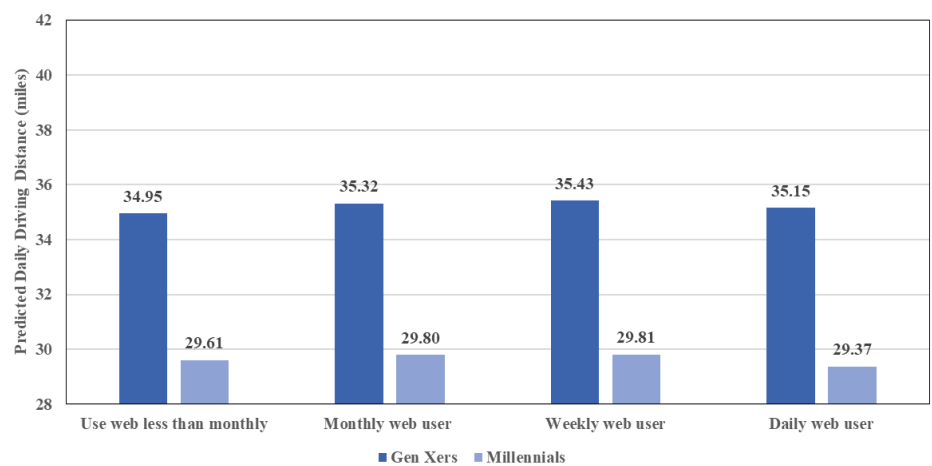

(b) internet usage

**Figure 3. The Marginal Effects of ICT**

*Relative contributions of life course, socioeconomic and other factors*

Although residential and economic factors remain the "largest contributors" of the variations of daily driving distances, life course events and ICT usages still account for nearly 15% of the total variations. Table 4 summarizes and ranks the relative contributions of independent variables on driving distance. The relative contributions are measured in percentage terms adding up to 100%. Among the explanatory variables, residential density is the largest contributor, with a relative contribution of 17.5%. The result confirms the claim that built environment could play a crucial role in influencing young adults' automobility (Ewing & Cervero, 2010; Stevens, 2017). The number of household vehicles per person, with a contribution of 12.1%, is the second most important variable in predicting young adults' driving distance. Annual household income is also a key predictor to daily driving distance with a relative contribution of 9.8%. This finding implies that higher income not only enables young adults to afford transportation but also allows them to participate in the various discretionary activities that require driving (Polzin et al., 2014).

Table 4. Relative Contributions of Independent Variables on Driving Distance

|  | Relative contribution (%) | Rank |
|---|---|---|
| **Residential location choice** (total contribution: 31.45%) | | |
| Population density (1,000 people per square mile) | 17.54 | 1 |
| States | 7.26 | 4 |
| Size of Metropolitan areas | 5.48 | 7 |
| Neighborhood type: urban | 1.18 | 18 |
| **Economic conditions** (total contribution: 21.96%) | | |
| Household vehicles per person | 12.12 | 2 |
| Household incomes (2017$, 1,000s) | 9.84 | 3 |
| **Life course events** (total contribution: 8.74%) | | |
| Has child under 15-year-old | 2.26 | 15 |
| Lives with parents or older relatives | 1.32 | 17 |
| Homeownership | 0.86 | 19 |
| Household size | 4.31 | 10 |
| **ICT usage** (total contribution: 5.00%) | | |
| Frequency of telecommute | 3.65 | 11 |
| Access to internet | 1.34 | 16 |
| **Demographics** (total contribution: 24.33%) | | |
| Age | 6.57 | 6 |
| Male | 6.95 | 5 |
| Education | 3.55 | 12 |
| Worker | 4.89 | 9 |
| Race | 2.37 | 14 |
| **Survey Years (2001 vs. 2017)** | 5.13 | 8 |
| **Weekend** | 3.38 | 13 |

If we cluster the variables into groups, residential location characteristics and economic conditions account for more than 50% of the total contributions to the young adults' daily driving distances. Using the more flexible data-driven method, this study confirms the argument from the regression-based literature that personal wealth and neighborhood design patterns play a critical role in shaping people's automobility (e.g., Brownstone and Golob, 2009; Ewing and Cervero, 2010). Also, the collective contribution of life course events and ICT usage is around 15%. This finding confirms the arguments that the advent of internet usage and delayed life course events can explain the reduced automobility among young adults (van Wee, 2015; Garikapati et al., 2016). While ICT usage and delayed lifecycle milestones clearly explain the young adults' driving distances, their effects are relatively smaller than built environment and personal wealth factors. Besides, demographic characteristics contribute about 25% of the total variations of young adults' driving distance. Finally, the categorical variable "survey year", a proxy for year-specific factors such as gasoline price trends, explains 5.1% of the variations of the dependent variable.

**Discussion**

By applying the GBDT models to analyze daily driving distance for the young adults in 2017 (Millennials) and 2001 (Gen Xers), this study investigates the association between young adults' automobility and various contributing factors without the linear-additive assumptions. The non-linear dose-response relationships for residential density, annual household incomes, and per capita number of vehicles have similar shapes for Millennials and Gen Xers, indicating similar non-linear relationships between these two generations. When in the same residential patterns, socioeconomic conditions, and lifecycle stages, Millennials' predicted driving distances are consistently lower than that of Gen Xers. This study also finds that residential and economic factors account for more than half of the total variations in daily driving distances; while the relative contributions of life course events and ICT usage are about 8.7% and 5.0%, respectively.

The non-linear dose-response effects of residential density, household incomes, and vehicle ownership complement the related regression-based studies in the literature (e.g., Blumenberg et al., 2016; McDonald, 2015). The regression-based studies found that these factors are significantly associated with the young adults' automobility, while this GDBT-based study advances from the regressions' average effects to identify the ranges that each variable has larger marginal effects (e.g., Ding et al., 2018a). Millennial young adults have consistently lower predicted driving distances than Gen Xers across different residential, economic, and lifecycle conditions. The findings are consistent with earlier empirical evidence based on regressions (e.g., Wang and Akar, 2020; Wang, 2019). Monthly and weekly telecommuters have relatively longer daily driving distances. One potential explanation would be the disruptive effect of teleworking on daily time travel allocation (e.g., Wang and Ozbilen, 2020); however, exploring the mechanisms of this effect is beyond the scope of this study. Finally, our estimations of the relative contributions of residential, economic, life course and demographic factors enrich the mainly regression-based literature covering the same topic (e.g., McDonald, 2015; Zhong and Lee, 2017).

Our findings imply that Millennials are less auto-centric than their proceeding generations, *ceteris paribus,* in both low-density and high-density neighborhoods. Planners and policymakers may foresee an opportunity to reshape the urban and suburban communities to be less auto-centric and more transit-friendly. Our study indicates that even if Millennial young adults will move from urban neighborhoods to suburban neighborhoods (Blumenberg et al., 2019), "catch up" their delayed lifecycles to form family and raise children (Delbosc and Nakanishi, 2017), and earn higher income when the economy recovers (Blumenberg et al., 2016), they might still be less car-reliant than their Gen-X counterparts (Lee et al., 2019a; Lee et al., 2019b). Admittedly, whether this generational difference will be truly consistent over time is still an open question. Although we do not have evidence to prove that Millennials will certainly keep driving less than Gen-Xers at an older age; at least at the age of 21-36, Millennials do show a lower level of automobility than Gen-Xers at the same age level when residing in neighborhoods with similar density levels (or the same urban or suburban neighborhoods). Based on the existing evidence, planners and policymakers should at least be aware of the possibility and be proactive in seeking out policy solutions, such as bundling mobility services into a Mobility as a Service (MaaS) product at different geographical contexts.

Particularly, we propose with caution that there might be an opportunity for the planners to make America's suburban communities less reliant on cars and more friendly to mass transit, in case suburban Millennials' current lower automobile dependent would persist over time. One key barrier preventing suburban residents from using transit is the first/last-mile problem, such as connecting transit stations with their homes (Boarnet et al., 2017). Planners and practitioners could seek for the first/last-mile connections by either providing ride-hailing service vouchers or working with high-tech companies to offer customized, crowd-sourced paratransit services. Furthermore, our non-linear dose-response plot shows that the marginal effects of residential density is the highest in the range of 0~6000 persons per square miles. In other words, land-use policies transforming highly sparse suburban neighborhoods into moderately sparse suburban neighborhoods (i.e., 5,000~6,000 persons per square mile) with higher supply of transit could sufficiently reduce the Millennial suburbanites' auto usage.

This study has the following limitations, many of which should motivate further research. First, as a data-driven, non-parametric approach, the GBDT models are not able to conduct statistical inferences such as significant tests or confidence intervals. GBDT models are also facing the risk of overfitting, users should be cautious when setting tuning parameters (e.g., tree complexity) and apply cross-validations (Tao et al., 2020). Second, GBDT models, like most other data-driven models, can only interpret the findings as associations as rather than causations (Mannering et al., 2020). A hybrid approach that combines machine learning algorithms and longitudinal data collection for behavioral changes may both uncover causality and achieve higher prediction accuracy (Athey, 2019). Third, although our findings show that the existing generational differences in automobility is more due to attitudes and preferences and less to residential patterns and life stages, this cannot guarantee that Millennials will definitely drive less than Gen-Xers at later stages of life, the Millennials' attitudes and preferences would change over time. Follow-up studies in the future would help to test on this. Fourth, the trip distances in 2001 and 2017 NHTSs are calculated using different methods. Although we have followed the Federal Highway Administration's guidelines to inflate the 2017 driving distances by 10%, there are still possibilities that the adjusted 2017 trip distances are not comparable with the 2001 ones. Fifth, although the study sample covers both urban and rural young adults, most of the discussion focuses on the "urban vs. suburban" dichotomy. More studies on this topic focusing on exurban or rural communities are needed as these geographic areas bear equal policy importance as urban and suburban areas.

**Conclusions**

Using national-representative travel survey data in 2001 and 2017, this study examines the factors associated with automobility for Millennial and Gen X young adults. Utilizing the gradient-boosting decision tree (GBDT) machine learning methods, this study relaxes the linear-additive assumptions for regression-based models for travel behaviors. Holding other factors constant, Millennial young adults have consistently lower expected daily driving distances than their Gen-X counterparts across different levels of residential patterns, economic conditions, lifecycle milestones and ICT usages. Comparing the relative contributions across different categories of factors, lifecycle events and ICT applications explains only 8.3% of the total variations of daily driving distances, while the residential and economic factors explain about 50% of the total variations. This study implies that as of 2017, Millennials were less auto-centric than their proceeding generations even taking residential patterns, economic conditions and lifecycle factors into considerations. Planners and policy makers should take this fact into account and be aware of the opportunities to transform America's suburbs to be less auto-centric and more transit-friendly for their Millennial dwellers. However, we acknowledge that the generational difference in young adulthood cannot be guaranteed to persist over later life stages. Therefore, this will be an important question to re-visit when the new round of nationwide travel surveys become available.

# References

Athey, S. (2019). The impact of machine learning on economics. In *The economics of artificial intelligence: An agenda* (pp. 507-547). University of Chicago Press.

Blumenberg, E., Brown, A., Ralph, K., Taylor, B. D., & Turley Voulgaris, C. (2019). A resurgence in urban living? Trends in residential location patterns of young and older adults since 2000. *Urban Geography*, *40*(9), 1375-1397.

Blumenberg, E., Ralph, K., Smart, M., & Taylor, B. D. (2016). Who knows about kids these days? Analyzing the determinants of youth and adult mobility in the US between 1990 and 2009. Transportation Research Part A: Policy and Practice, 93, 39-54.

Boarnet, M. G., Giuliano, G., Hou, Y., & Shin, E. J. (2017). First/last mile transit access as an equity planning issue. *Transportation Research Part A: Policy and Practice*, *103*, 296-310.

Boarnet, M. G., Wang, X., & Houston, D. (2017). Can new light rail reduce personal vehicle carbon emissions? A before‐after, experimental‐control evaluation in Los Angeles. *Journal of Regional Science*, *57*(3), 523-539.

Breiman, L., Friedman, J., Stone, C. J., & Olshen, R. A. (1984). Classification and regression trees. CRC Press.

Brownstone, D., & Golob, T. F. (2009). The impact of residential density on vehicle usage and energy consumption. Journal of Urban Economics, 65(1), 91.

Burstein, David D. (2013). Fast Future: How the Millennial Generation is Shaping Our World. Boston: Beacon Press

Choi, K., Jiao, J., & Zhang, M. (2017). Reducing Vehicle Travel for the Next Generation: Lessons from the 2001 and 2009 National Household Travel Surveys. Journal of Urban Planning and Development, 143(4), 04017017.

Chung, Y. S. (2013). Factor complexity of crash occurrence: An empirical demonstration using boosted regression trees. Accident Analysis & Prevention, 61, 107-118.

Circella, G., Fulton, L., Alemi, F., Berliner, R. M., Tiedeman, K., Mokhtarian, P. L., & Handy, S. What Affects Millennials' Mobility? PART I: Investigating the Environmental Concerns, Lifestyles, Mobility-Related Attitudes and Adoption of Technology of Young Adults in California. National Center for Sustainable Transportation, University of California, Davis, CA, 2016

Claritas. (2018). Assessing the Role of Urbanicity. Retrieved from https://nhts.ornl.gov/assets/Assessing_the_Role_of_Urbanicity.pdf

da Silva, D. C., Astroza, S., Batur, I., Khoeini, S., Magassy, T. B., Pendyala, R. M., & Bhat, C. R. (2019). Are Millennials Really All That Different Than Generation X? An Analysis of Factors Contributing to Differences in Vehicle Miles of Travel.

Delbosc, A., & Currie, G. (2013). Causes of youth licensing decline: a synthesis of evidence. Transport Reviews, 33(3), 271-290.

Delbosc, A., & Nakanishi, H. (2017). A life course perspective on the travel of Australian millennials. *Transportation Research Part A: Policy and Practice, 104*, 319-336.

Delbosc, A., & Ralph, K. (2017). A tale of two millennials. Journal of Transport and Land Use, 10(1), 903-910.

Delbosc, A., McDonald, N., Stokes, G., Lucas, K., Circella, G., & Lee, Y. (2019). Millennials in cities: Comparing travel behaviour trends across six case study regions. Cities, 90, 1-14.

DePillis, L. (2017). 10 years after the recession began, have Americans recovered? Retrieved from https://money.cnn.com/2017/12/01/news/economy/recession-anniversary/index.html

DeVaney, S. A. (2015). Understanding the millennial generation. Journal of Financial Service Professionals, 69(6).

Ding, C., Cao, X. J., & Næss, P. (2018a). Applying gradient boosting decision trees to examine non-linear effects of the built environment on driving distance in Oslo. Transportation Research Part A: Policy and Practice, 110, 107-117.

Ding, C., Chen, P., & Jiao, J. (2018b). Non-linear effects of the built environment on automobile-involved pedestrian crash frequency: a machine learning approach. Accident Analysis & Prevention, 112, 116-126.

Dong, W., Cao, X., Wu, X., & Dong, Y. (2019). Examining pedestrian satisfaction in gated and open communities: An integration of gradient boosting decision trees and impact-asymmetry analysis. Landscape and Urban Planning, 185, 246-257.

Ewing, R., & Cervero, R. (2010). Travel and the built environment: a meta-analysis. Journal of the American planning association, 76(3), 265-294.

Federal Highway Administration, (2004). 2001 National Household Travel Survey User's Guide. Retrieved from http://nhts.ornl.gov/2001/usersguide/UsersGuide.pdf.

Federal Highway Administration, (2018a). 2017 National Household Travel Survey User's Guide. Retrieved from https://nhts.ornl.gov/assets/2017UsersGuide.pdf.

Federal Highway Administration (2018b). 2017 NHTS Technical Release Notes. Retrieved from https://nhts.ornl.gov/assets/2017_NHTS_Technical_Release_Notes_030818sb_revised112018.pdf

Frey, W. (Producer). (2018, January, 2019). Millennials are the largest generation in theU.S. labor force. Retrieved from http://www.pewresearch.org/facttank/2018/04/11/millennials-largest-generation-us-labor-force/

Friedman, J. H. (2001). Greedy function approximation: a gradient boosting machine. Annals of statistics, 1189-1232.

Furman, Jason. 2014. "America's Millennials in the Recovery." Keynote address at the Zillow Housing Forum, Washington, DC, July 24. Accessed June 28, 2018.

Garikapati, V. M., Pendyala, R. M., Morris, E. A., Mokhtarian, P. L., & McDonald, N. (2016). Activity patterns, time use, and travel of millennials: a generation in transition?. Transport Reviews, 36(5), 558-584.

Greenwell, B. (2018). Partial Dependence Plots [R package pdp version 0.7.0]. Retrieved from https://cran.r-project.org/web/packages/pdp/

Greenwell, B., Boehmke, B., Cunnningham, J., Developers, GBM (2019). Generalized Boosted Regression Models. R package gbm version 2.1.5. Retrieved from https://cran.r-project.org/web/packages/gbm/

Hastie, T., Tibshirani, R., & Friedman, J. (2009). The elements of statistical learning: Data mining, inference and prediction, second edition. Springer, New York.

Hjorthol, R. (2016). Decreasing popularity of the car? Changes in driving licence and access to a car among young adults over a 25-year period in Norway. Journal of transport geography, 51, 140-146.

Hopkins, D., 2016. Can environmental awareness explain declining preference for car-based mobility amongst generation Y? A qualitative examination of learn to drive behaviours. Transport. Res. Part A: Policy Practice 94, 149–163.

Klein, N. J., & Smart, M. J. (2017). Millennials and car ownership: Less money, fewer cars. Transport Policy, 53, 20-29.

Knittel, C. R., & Murphy, E. (2019). Generational trends in vehicle ownership and use: Are millennials any different? (No. w25674). National Bureau of Economic Research.

Kuhnimhof, T., Armoogum, J., Buehler, R., Dargay, J., Denstadli, J. M., & Yamamoto, T. (2012). Men shape a downward trend in car use among young adults—evidence from six industrialized countries. Transport Reviews, 32(6), 761-779.

Lee, H. (2018). Are Millennials Coming to Town? Residential Location Choice of Young Adults. Urban Affairs Review, 1078087418787668.

Lee, Y., & Circella, G. (2019). ICT, millennials' lifestyles and travel choices. The Evolving Impact of ICT on Activities and Travel Behaviour, 3, 107.

Lee, Y., Circella, G., Mokhtarian, P. L., & Guhathakurta, S. (2019a). Are millennials more multimodal? A latent-class cluster analysis with attitudes and preferences among millennial and Generation X commuters in California. Transportation, 1-24.

Lee, Y., Circella, G., Mokhtarian, P. L., & Guhathakurta, S. (2019b). Heterogeneous residential preferences among millennials and members of generation X in California: A latent-class approach. Transportation Research Part D: Transport and Environment, 76, 289-304.

Luhby, T (2018). Millennials born in the 1980s may never recover from the Great Recession. Retrieved from https://money.cnn.com/2018/05/22/news/economy/1980s-millennials-great-recession-study/index.html

Ma, X., Ding, C., Luan, S., Wang, Y., Wang, Y., 2017. Prioritizing influential factors for freeway incident clearance time prediction using the gradient boosting decision trees method. IEEE Trans. Intell. Transp. Syst. 18, 2303–2310.

Ma, X., Wu, Y. J., Wang, Y., Chen, F., & Liu, J. (2013). Mining smart card data for transit riders' travel patterns. Transportation Research Part C: Emerging Technologies, 36, 1-12.

Mannering, F., Bhat, C. R., Shankar, V., & Abdel-Aty, M. (2020). Big data, traditional data and the tradeoffs between prediction and causality in highway-safety analysis. Analytic methods in accident research, 25, 100113.

Manville, M., King, D. A., & Smart, M. J. (2017). The driving downturn: a preliminary assessment. Journal of the American Planning Association, 83(1), 42-55.

Mawhorter, Sarah L. 2018. "Boomers and Their Boomerang Kids: Comparing Housing Opportunities for Baby Boomers and Millennials in the United States." In The Millennial City: Trends, Implications, and Prospects for Urban Planning and Policy, edited by Markus Moos, Deirdre Pfeiffer, and Tara Vinodrai, 143–52. New York: Routledge.

McDonald, N. C. (2015). Are millennials really the "go-nowhere" generation?. Journal of the American Planning Association, 81(2), 90-103.

McDonald, N. C., Merlin, L., Hu, H., Shih, J., Cohen, D. A., Evenson, K. R., McKenzie, T. L., & Rodriguez, D. A. (2016). Longitudinal analysis of adolescent girls' activity patterns: Understanding the influence of the transition to licensure. Journal of transport and land use, 9(2), 67.

Myers, D. (2016). Peak millennials: Three reinforcing cycles that amplify the rise and fall of urban concentration by millennials. Housing Policy Debate, 26(6), 928-947.

Pirra, M., & Diana, M. (2019). A study of tour-based mode choice based on a Support Vector Machine classifier. Transportation Planning and Technology, 42(1), 23-36.

Polzin, S. E., Chu, X., & Godfrey, J. (2014). The impact of millennials' travel behavior on future personal vehicle travel. Energy Strategy Reviews, 5, 59-65.

Ralph, K. M. (2017). Multimodal millennials? The four traveler types of young people in the United States in 2009. Journal of Planning Education and Research, 37(2), 150-163.

Saha, D., Alluri, P., & Gan, A. (2015). Prioritizing Highway Safety Manual's crash prediction variables using boosted regression trees. Accident Analysis & Prevention, 79, 133-144.

Scheiner, J., & Holz-Rau, C. (2013). A comprehensive study of life course, cohort, and period effects on changes in travel mode use. Transportation Research Part A: Policy and Practice, 47, 167-181.

Smart, M. J., & Klein, N. J. (2018). Remembrance of cars and buses past: how prior life experiences influence travel. *Journal of Planning Education and Research*, *38*(2), 139-151.

Susilo, Y. O., Liu, C., & Börjesson, M. (2019). The changes of activity-travel participation across gender, lifecycle, and generations in Sweden over 30 years. Transportation, 46(3), 793-818.

Stevens, M. R. (2017). Does compact development make people drive less?. Journal of the American Planning Association, 83(1), 7-18.

Talen, E. (2017). Empower the Millennials. Housing Policy Debate, 27(2), 331-333.

Tao, T., Wang, J., & Cao, X. (2020). Exploring the non-linear associations between spatial attributes and walking distance to transit. Journal of Transport Geography, 82, 102560.

Thigpen, C., & Handy, S. (2018). Driver's licensing delay: A retrospective case study of the impact of attitudes, parental and social influences, and intergenerational differences. Transportation research part A: policy and practice, 111, 24-40.

U.S. Census Bureau (2019). About 13.1 Percent Have a Master's, Professional Degree or Doctorate. Retrieved from: https://www.census.gov/library/stories/2019/02/number-of-people-with-masters-and-phd-degrees-double-since-2000.html

van Wee, B. (2015). Peak car: The first signs of a shift towards ICT-based activities replacing travel? A discussion paper. Transport Policy, 42, 1-3.

Wang, K., Akar, G., & Chen, Y. J. (2018). Bike sharing differences among Millennials, Gen Xers, and Baby Boomers: Lessons learnt from New York City's bike share. Transportation Research Part A: Policy and Practice, 116, 1-14.

Wang, X. (2019). Has the relationship between urban and suburban automobile travel changed across generations? Comparing Millennials and Generation Xers in the United States. *Transportation Research Part A: Policy and Practice, 129*, 107-122.

Wang, K., & Akar, G. (2020). Will Millennials Drive Less as the Economy Recovers: A Postrecession Analysis of Automobile Travel Patterns. Journal of Planning Education and Research, 0739456X20911705.

Wang, K., & Ozbilen, B. (2020). Synergistic and threshold effects of telework and residential location choice on travel time allocation. *Sustainable Cities and Society, 63*, 102468.

Winship, C., & Radbill, L. (1994). Sampling weights and regression analysis. *Sociological Methods & Research, 23*(2), 230-257.

Xie, C., Lu, J., & Parkany, E. (2003). Work travel mode choice modeling with data mining: decision trees and neural networks. Transportation Research Record, 1854(1), 50-61.

Zhang, Y., & Haghani, A. (2015). A gradient boosting method to improve travel time prediction. Transportation Research Part C: Emerging Technologies, 58, 308-324.

Zhong, L., & Lee, B. (2017). Carless or car later?: Declining car ownership of millennial households in the Puget Sound Region, Washington State. Transportation Research Record, 2664(1), 69-78.